\documentclass[10pt,twocolumn,letterpaper]{article}

\usepackage{cvpr}
\usepackage{times}
\usepackage{epsfig}
\usepackage{graphicx}
\usepackage{amsmath}
\usepackage{amssymb}
\usepackage{bm}

\usepackage[breaklinks=true,bookmarks=false]{hyperref}

\newcommand{\PutCaptrain}{\put(2,3){{~~~~~~~~~~~~~~~~~~~~~~~~~~~~~~~~~~~~~~~~~~~~~~~~~~(a)~~~~~~~~~~~~~~~~~~~
~~~~~~~~~~~~~~~~~~~~~~~~~~~~~~~~~~~~~~~~~~~~~~~~~~~~~~~~~~~~~~~~~~~~~~(b)~~~~~~~~~~~~~~~~~~ }}}

\cvprfinalcopy 

\def\cvprPaperID{****} 

\ifcvprfinal\fi
\begin{document}

\title{Learning Spatial-Aware Regressions for Visual Tracking}

\author{Chong Sun$^{1,2}$, Dong Wang$^1$, Huchuan Lu$^1$, Ming-Hsuan Yang$^2$\\
$^1$School of Information and Communication Engineering, Dalian University of Technology, China\\ $^2$Electrical Engineering and Computer Science, University of California, Merced, USA\\
{\tt\small waynecool@mail.dlut.edu.cn, \{wdice,lhchuan\}@dlut.edu.cn,  mhyang@ucmerced.edu}
}

\maketitle
\thispagestyle{empty}

\begin{abstract}
 In this paper, we analyze the spatial information of deep features, and propose two complementary
 regressions for robust visual tracking.
 First, we propose a kernelized ridge regression model wherein the kernel value is defined as the
 weighted sum of similarity scores of all pairs of patches between two samples.
We show that this model can be formulated as a neural network and thus can be efficiently solved.
Second, we propose a fully convolutional neural network with spatially regularized kernels, through
which the filter kernel corresponding to each output channel is forced to focus on a specific region of the target.
Distance transform pooling is further exploited to determine the effectiveness
of each output channel of the convolution layer.
The outputs from the kernelized ridge regression model and the fully convolutional neural network
are combined to obtain the ultimate response.
Experimental results on two benchmark datasets validate the effectiveness of the proposed method.

\end{abstract}

\begin{figure}[t]
\centering
\begin{tabular}{c@{}c}

\includegraphics[width=0.48\linewidth,height=20mm]{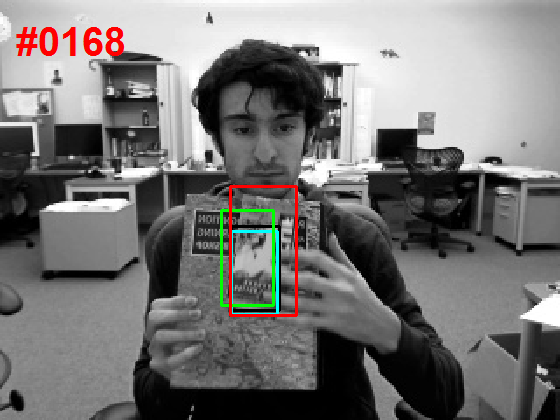}
\ &
\includegraphics[width=0.48\linewidth,height=20mm]{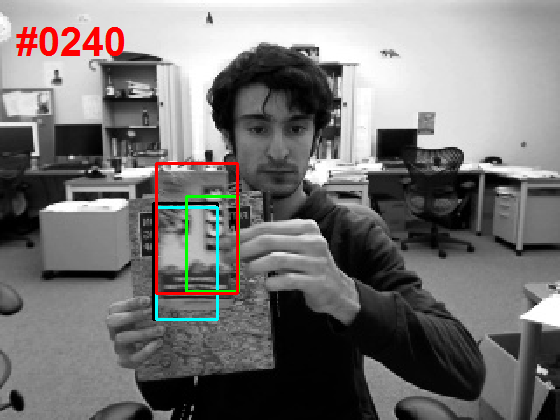}
\ \\
\includegraphics[width=0.48\linewidth,height=20mm]{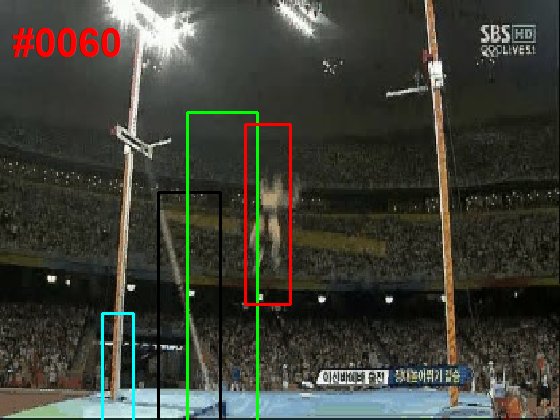}
\ &
\includegraphics[width=0.48\linewidth,height=20mm]{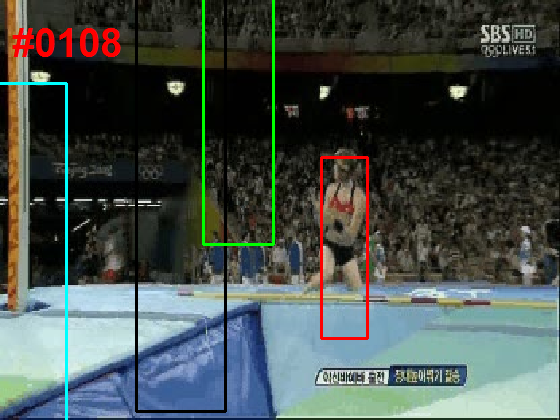}
\ \\
\includegraphics[width=0.48\linewidth,height=20mm]{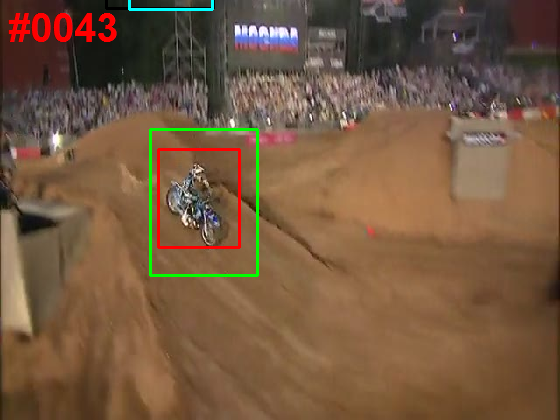}
\ &
\includegraphics[width=0.48\linewidth,height=20mm]{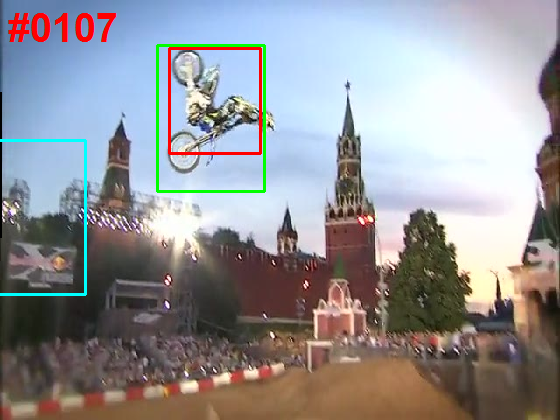}
\ \\
\end{tabular}
\begin{tabular}{c}
{\kern 6mm}\includegraphics[width=0.7\linewidth]{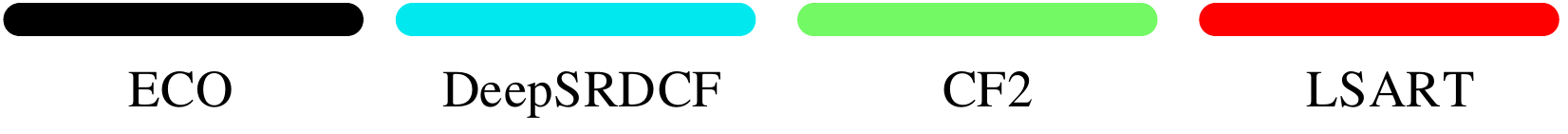}
\end{tabular}
\caption{Example tracking results of different methods on the OTB dataset. Our tracker (LSART) has comparable results compared with the state-of-the-art algorithms.}
\label{fig:tracking_results}
\vspace{-2mm}
\end{figure}

\section{Introduction}
Visual tracking, which aims to continuously estimate the positions and scales of a
pre-specified target, has been a hot topic for the last decades. It is widely
used in numerous vision tasks, such as video surveillance, augmented reality and so on.
Current algorithms have achieved very impressive results, however, many problems
remain to be solved.

With the emergence of large-scale datasets, deep neural networks have shown their
great capacity in object classification, image identification, to name a few.
It has been verified in many prior papers~\cite{wang2015visual,nam2016learning,LiWWL18}
that trackers based on convolutional neural networks (CNNs) can significantly
improve the tracking performance. Usually, these methods pretrain their networks
on a large scale dataset, and finetune the networks with the ground-truth data
in the first frame of a sequence.
In addition to the CNN-based trackers, methods based on the kernelized correlation
filter (KCF) are also very popular in recent years for the efficiency and capacity
to utilize large numbers of negative samples. As is described in~\cite{henriques2015high},
the KCF method is essentially the kernelized ridge regression (KRR) with cyclically
shifted samples. Methods based on the KCF usually take a region of interest as the
input, which makes it very difficult to exploit the structural information of the
target. In addition, the cyclically constructed samples also introduce the unwanted
boundary effects. Compared to the KCF method, the dominating reason why the conventional
KRR is not widely applied is that it has to compute a kernel matrix for large numbers of
samples, which results in heavy computational load.

Both the CNN-and KRR-based trackers have limitations and they have complementarities.
The CNN-based trackers usually contain a large number of parameters which are difficult
to be finetuned in the tracking problem. As a result, the trained filter kernels in
the convolution layer are usually highly correlated and tend to overfit the training data.
On the contrary, the KRR-based trackers have limited model parameters (equal to the number
of training samples), and cannot learn discriminative enough models when training samples
are correlated. In addition, the existing KRR-based methods assume that each part of the
target is equally important, and ignore the relationship among different parts.

In this paper, we exploit both the KRR and CNN models,
and learn the complementary spatial-aware regressions for visual tracking (LSART).
First, we propose a kernelized ridge regression with cross-patch similarities.
We assign each similarity score a weight, and simultaneously learn this weight and
ridge regression model parameters.
We show that the proposed ridge regression model can be reformulated as a neural network,
which is more efficiently optimized than the original form.
Second, we introduce the spatially regularized kernels into the fully convolutional
neural network. By imposing spatial constraints on the filter kernels, we enforce each
output channel of the convolution layer to have a response for a specific localized region.
We exploit the distance transform pooling layer to determine the effectiveness of the
outputs from the convolution layer, and develop a two-stage training strategy to update
the CNN model effectively.
Finally, the heat maps obtained by the KRR and CNN models are combined to generate
a final heat map for target location.
Experiments on the popular datasets demonstrate that our tracker performs significantly
better than other state-of-the-art methods (see Figure~\ref{fig:tracking_results} for visualized tracking results).

The main contributions of this paper can be summarized as follows:

1. We develop a spatial-aware KRR model by introducing a cross-patch
similarity kernel. This model can model both regression coefficients
and patch reliability, which enables our model to be robust to the unreliable patches.
The regression coefficient and similarity weight vectors are simultaneously
optimized via an end-to-end neural network, which is new in visual tracking
and facilitates seamlessly integrating our model with deep feature extraction
networks.

2. We propose the spatially regularized filter kernels for CNN, which enforces
each filter kernel to focus on a localized region.
We also design a two-stream training network to effectively learn network
parameters, which avoids overfitting and considers rotation information.

3. We propose the complementary KRR and CNN models based on their inherent
limitations. The spatial-aware KRR model focuses on the holistic object and
the spatial-aware CNN model focuses on small and localized regions, thereby
complementing each other for better performance.

4. Our method achieves very promising tracking performance, especially
on the recent VOT-2017 public dataset.

\section{Related Work}
Algorithms of visual tracking mainly focus on designing robust appearance
models, which are roughly categorized as generative and discriminative models.
With the progress of deep neural networks and correlation filters, discriminative
appearance models are preferred in the recent works.

Trackers based on correlation filter have attracted more and more attention for
the advantages in efficiency and robustness. In essential, correlation filter can
be viewed as a kernelized ridge regression (KRR) model that can be speeded up in
the frequency domain.
Bolme \etal~\cite{bolme2010visual} exploit the correlation filter with minimum
output sum of squared error for visual tracking. As fast Fourier transform is used,
the tracker achieves very fast tracking performance.
In~\cite{henriques2012exploiting}, Henriques~\etal first incorporate kernel functions
into the correlation filter, which is named as CSK. The CSK tracker can also be solved
via fast Fourier transform, thus is efficient.
Based on~\cite{henriques2012exploiting}, the tracking method~\cite{henriques2015high}
further improves the CSK tracker by using the histogram of gradient features.
Ma~\etal~\cite{ma2015hierarchical} exploit complementary nature of features extracted
from three layers of CNN, and use a coarse-to-fine scheme for target searching.
Based on~\cite{ma2015hierarchical}, an online adaptive Hedge method~\cite{qi2016hedged}
is designed, which takes both the short-term and long-term regrets into consideration.
In this method, they use the CF-based tracker defined on a single CNN layer as an expert
and learn the adaptive weights for different experts.
Danelljan~\etal propose several CF-based trackers with good performance.
The SRDCF method~\cite{danelljan2015learning} tries to suppress the boundary
effects of the correlation filter by multiplying the filter coefficients with
spatial regularization weights produced by a Gaussian distribution.
This tracker achieves very good performance even with hand-crafted features.
Based on~\cite{danelljan2015learning}, they propose an adaptive decontamination
method~\cite{danelljan2016adaptive} for the correlation filter, which adaptively
learns the reliability of each training sample and eliminates the influence of
contaminated ones.
Furthermore, the learning process for correlation filter is conducted in the
continuous spatial domain of various feature maps~\cite{danelljan2016beyond},
which incorporates the sub-pixel information.
These methods usually exploit the linear or Gaussian kernel to depict the
similarity between the target region and a given candidate region.
They inevitably ignore the intrinsic structure information within the target
region, which makes the trackers be less effective in dealing with occlusion
and deformation challenges.
In this work, we introduce a novel kernel to model the cross-patch similarities,
develop the corresponding KRR optimization model, and provide a network structure
to solve it efficiently.

Compared with trackers based on correlation filters, CNN-based trackers also
achieve good performance in recent years. In~\cite{li2014robust}, a shallow network
with two convolution layers is proposed, which learns the feature representation
and classifier simultaneously. In~\cite{wang2015visual}, Wang~\etal transfer the
model pre-trained on image classification dataset to visual tracking and exploit
the fully convolutional neural network for target location.
Wang~\etal~\cite{wang2016stct} propose a sequentially training fashion for neural
network to avoid the overfitting problem. Tao~\etal~\cite{tao2016siamese} off-line
train a Siamese deep neural network on large amounts of extra video sequences, and
directly apply this model to conduct the optimal match during the tracking process.
In~\cite{nam2016learning}, a multi-domain CNN-based tracker is proposed,
in which the shared layers are used to obtain generic target representation
and the domain specific layers are adopted for classification.
These CNN-based trackers usually exploit the global information regarding the target
object and therefore ignore its spatial layouts. To address this issue, we propose
a convolution layer with spatially regularized filter kernels to focus on local
regions of the target. Besides, we effectively combine the proposed KRR and CNN models
to develop a robust tracker.

\section{Spatial-Aware KRR}
\subsection{KRR with Cross-Patch Similarity (KRRCPS)}
Given $N$ training samples $\{ ({{\bf x}_i},{y_i})\} _{i = 1}^N$ in frame $t$,
the conventional ridge regression can be formulated as
\begin{equation}
{\bf w}_t=\arg\min_{{\bf w}_t} \sum\limits_i {{{({y_i} - {{\bf{w}}_t^\top}
{{\bf{x}}_i})}^2} + \lambda {{\left\| {\bf{w}}_t \right\|}^2}},
\label{eq:3-1}
\end{equation}
where ${{\bf x}_i}\in \mathbb{R}^{d\times 1}$ denotes the feature vector for sample $i$,
and $y_i$ is its sample label.
In Eq.~\ref{eq:3-1}, ${\bf w}_t\in \mathbb{R}^{d\times 1}$ denotes the
linear weight vector, and $\lambda$ is a regularization constant.
According to the representer theorem~\cite{representertheorem}, the model parameter
${\bf w}_t$ can be determined as the weighted sum of the training samples (i.e.,
${\bf{w}}_t = \sum\limits_{i = 1}^N {{\alpha _i}^t{{\bf{x}}_i}} $).
Thus, the optimization problem (\ref{eq:3-1}) can be rewritten as
\begin{equation}
{\bm \alpha}_t^{*}=\arg\min_{ {\bm \alpha}_t} \sum\limits_{i=1}^N {{{({y_i}
- \sum\limits_{j = 1}^N {{\alpha _j^t}{k_{ij}}} )^2}} +
\lambda } \sum\limits_{i, j=1}^N {\alpha _i^t\alpha_j^t{k_{i,j}}},
\label{eq:3-2}
\end{equation}
where $\alpha_i^t$ is the weight for sample $i$, ${\bm \alpha}_t=\left[\alpha_1^t,...,\alpha_N^t\right]^\top \in \mathbb{R}^{N\times 1}$,
and $k_{i,j}$ is the kernel value computed between features ${\bf x}_i$ and ${\bf x}_j$.
The existing kernel definitions (\eg, linear kernel and Gaussian kernel) do not fully
consider the spatial layouts of the target, which limits the tracking performance.

In this work, we introduce a kernel function which considers the similarity of all pairs of patches between two samples.
Especially, we divide each sample $i$ into $M$ patches, and obtain features for each patch $m$ as ${\bf x}_i^m$,
then the kernel value between sample $i$ and $j$ is computed as
\begin{equation}
{k_{ij}} = \sum\limits_{m, n = 1}^M {{\beta _{m,n}^t}{{{\bf{x}}_{i}^m}^\top}} {\bf{x}}_{j}^{n},
\label{eq:3-3}
\end{equation}
where $\beta_{m,n}^t$ denotes the weight for the similarity score between $m$-and $n$-th patches.
This kernel function has at least two advantages: (a) for each similarity score, the kernel function assigns a learnable
weight to make the model adaptively focus on the similarity scores of reliable regions; (b) more similarity pairs
between patches are considered, thereby enhancing the discriminant ability of the model.

By substituting Eq.\ref{eq:3-3} into Eq.\ref{eq:3-2} and introducing a regularization term, we obtain the
optimization problem (\ref{eq:optkrr}).
\begin{equation}
{\bm{\alpha }}_t^*,{\bm{\beta }}_t^* = \arg \mathop {\min }\limits_{{{\bm{\alpha }}_t},
{{\bm{\beta }}_t}} J\left( {{{\bm{\alpha }}_t},{{\bm{\beta}}_t}} \right),
\label{eq:optkrr}
\end{equation}
where we define $J\left( {{\bm{\alpha }},{\bm{\beta }}} \right)$ as
\begin{equation}
\begin{gathered}
J\left( {{\bm{\alpha }},{\bm{\beta }}} \right) \hfill \\
   = \sum\limits_{i = 1}^N {({y_i} - \sum\limits_{j = 1}^N {\alpha _j^t\sum\limits_{m, n = 1}^M
     {\beta _{m,n}^t{\mathbf{x}}{{_j^m}^ \top }{\mathbf{x}}_i^n} {)^2}} }  \hfill \\
     \;\;\;\; + {\lambda _1}\sum\limits_{i, j = 1}^N {\alpha _i^t} {k_{i,j}}\alpha _j^t +
     {\lambda _2}\sum\limits_{m, n = 1}^M {{{\left( {\beta _{m,n}^t} \right)}^2}}  \hfill \\
   = \left\| {{\mathbf{y}} - \sum\limits_{m, n = 1}^M {{\mathbf{f}}_m^ \top \beta _{m,n}^t
     {{\mathbf{f}}_n}}{{\bm{\alpha }}_t}} \right\|_2^2 + {\lambda _1}{\bm{\alpha }}_t^ \top
     {\mathbf{K}}{{\bm{\alpha }}_t} + {\lambda _2}\left\| {{{\bm{\beta }}_t}} \right\|_2^2, \hfill \\
\end{gathered}
\label{eq:objkrr}
\end{equation}
where ${\bm{\beta}}_t=\left[\beta_{1,1}^t,\beta_{1,2}^t,...,\beta_{M,M}^t\right]$ is the
weight vector for all cross patches, ${{\bf{f}}_m} = [{\bf{x}}_1^m,...,{\bf{x}}_N^m]$
stands for the concatenated feature matrix for the $m$-th patch of $N$ samples, and
$\bf K$ denotes the kernel matrix whose $(i,j)$-th element is ${k_{ij}} = \sum\limits_{m ,n = 1}^M
{{\beta _{m,n}^t}{{{\bf{x}}_{i}^m}^\top}} {\bf{x}}_{j}^n$.

A conventional solver for the optimization problem (\ref{eq:optkrr}) is the alternating
iteration algorithm that optimizes ${\bm{\alpha}}_t$ and ${\bm{\beta}}_t$ iteratively. If
${\bm{\beta}}_t$ is fixed, the analytical solution for ${\bm \alpha}_t$ can be obtained as
$\begin{array}{l}
{\bm{\alpha }}_t = {({\bf{K}} + {\lambda _1}{\bf{I}})^{ - 1}}{\bf{y}}\\
\end{array}$  ($\mathbf{I}$ denotes an identity matrix), whose computation complexity is $\mathcal{O}(N^3)$.
If ${\bm{\alpha}}_t$ is fixed, ${\bm \beta}_t$ can be updated via the gradient descent algorithm,
and it is easy to know that the computation complexity for this process is $\mathcal{O}(dN^2)$.
Thus, optimizing ${\bm{\alpha}}_t$ and ${\bm{\beta}}_t$ via the alternating
iteration algorithm is very time consuming for the online update process.

\begin{figure}[t]
\centering
\includegraphics[width=1\linewidth]{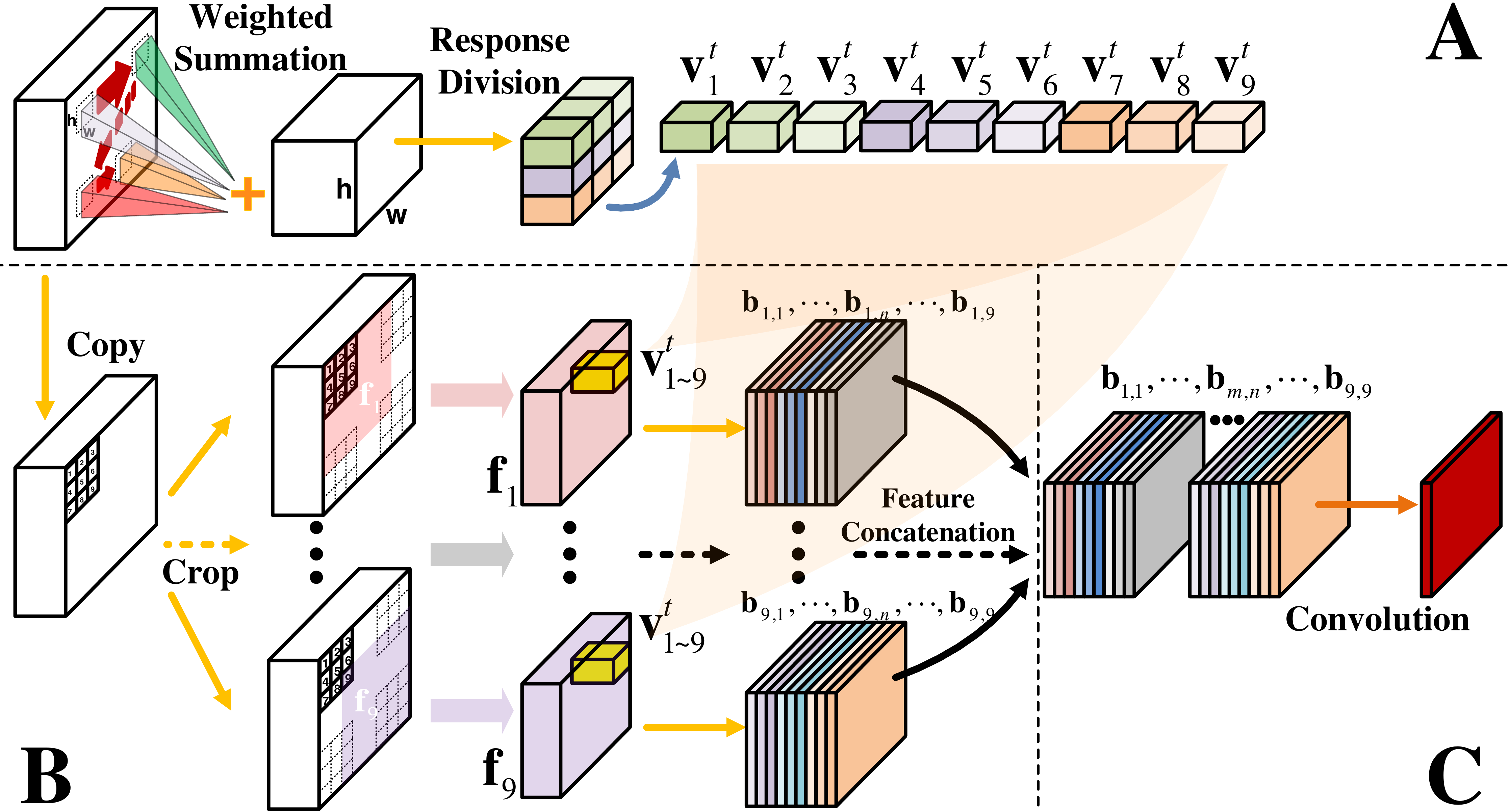}\\
\caption{The reformulated network structure for our KRRCPS model. In Module B, we show
two examples on how to crop the feature maps for ${\bf f}_1$ and ${\bf f}_9$.
By doing this, we make the responses generated by different fragments correspond to the same
 input ROI region.
  \textbf{Best viewed in high resolutions with
  zoom in.}}
  \label{fig:3-1}
  \vspace{-2mm}
\end{figure}

\subsection{Network Structure for KRRCPS}
In this work, we attempt to learn ${\bm \alpha}_t$ and ${\bm \beta}_t$ by reformulating
the proposed ridge regression into a neural network. This reformulation not only provides
an efficient solver but also enables the algorithm to be seamlessly integrated with
state-of-the-art deep feature extraction networks.

In Eq.~\ref{eq:objkrr}, the response term ${\bf r}=\sum\limits_{m ,n = 1}^M {{\bf{f}}_m^
\top{\beta _{m,n}^t}{{\bf{f}}_n}} {\bm{\alpha }}_t$ can be sequentially calculated by the
following three steps:
\begin{equation}
\begin{small}
\begin{array}{l}
{\mathbf{A:}}{\kern 12.0pt}{{\bf{v}}_n^t} = {{\bf{f}}_n}{\bm{\alpha }}_t\\
{\mathbf{B:}}{\kern 13.0pt}{{\bf{b}}_{m,n}} = {\bf{f}}_m^\top{{\bf{v}}_n^t}\\
{\mathbf{C:}}{\kern 15.0pt}{\bf r} = \sum\limits_{m, n = 1}^M {{\beta _{m,n}^t}}
{{\bf{b}}_{m,n}}
\end{array}.
\label{eq:3-7}
\end{small}
\end{equation}
Thus, we develop an equivalent neural network of our regression model (shown in
Figure~\ref{fig:3-1}), which takes the extracted feature map as the input and
outputs a heat map for target localization. This network consists of three modules,
each of which precisely corresponds to one of the three operations in Eq.~\ref{eq:3-7}.

\noindent{\textbf{Module A: }}Given the target location, we first crop a rectangle region around the target object,
and obtain the feature map ${\bf X}_t$ with size
$H\times W\times C$, thus the target size projected on the feature map is $h\times w$. Based on the projected target size, we densely
crop samples and reshape each sample to a $d$ ($=h\times w\times C$) dimensional vector.
This results in a feature matrix ${\bf D}_t\in \mathbb{R}^{d\times N}$, based on which the
output of the weighted sum layer is obtained as
\begin{small}
\begin{equation}
{\bf z}={\bf D}_t{\bm \alpha}_t,
\end{equation}
\end{small}
where ${\bf z}\in \mathbb{R}^{d\times 1}$ denotes the response for the current layer and
${\bm \alpha}_t$ is the weight vector to be learned. We reshape the vector ${\bf z}$ to
a response map with size $h\times w\times C$, and then divide it into $M=\sqrt{M}\times \sqrt{M}$
sub-responses with size $(h/\sqrt{M})\times (w/\sqrt{M})\times C$. We use ${\bf v}_1^t...{\bf v}_M^t$
to denote these sub-responses in Figure~\ref{fig:3-1}.

\vspace{1mm}
\noindent{\textbf{Module B: }}This module corresponds to the second equation of Eq.~\ref{eq:3-7},
which is equivalent to a convolution layer.
Noticing that the feature map corresponding to ${\bf f}_m, m\in\{1,...,M\}$ is a sub-region of
the input feature map ${\bf X}_t$ and is different when $m$ varies, we
first obtain the feature
map corresponding to ${\bf f}_m$ by cropping ${\bf X}_t$ (see \textbf{Module B} in
Figure~\ref{fig:3-1} for example) and feed it into a convolution
layer which takes ${\bf v}_1^t...{\bf v}_M^t$ as an ensemble of filter kernels.
As we have $M$ patches in total, the convolution layer has $M$ outputs, each of which has
the size $(H-h+1)\times (W-w+1) \times M$.

\vspace{1mm}
\noindent{\textbf{Module C: }}We concatenate the $M$ outputs of \textbf{Module B} through the
Concat layer, and input the concatenated feature maps into a convolution layer, whose kernel
size is $1\times 1 \times M^2$. This module corresponds to operation \textbf{C} in Eq.~\ref{eq:3-7},
and the filter kernel corresponds to ${\bm \beta}_t$.

In this work, we use the backpropagation algorithm to solve this network. The computational complexity
for both forward and backward propagations is $\mathcal{O}(dN)$, which is much more efficient than the
original solver.
Note that our network is defined based on Eq.~\ref{eq:objkrr} for model learning.
At the detection stage in frame $t$, we just need to replace ${\bf D}_t$, ${\bm \alpha}_t$ and
${\bm \beta}_t$ in the network with ${\bf \hat D}_t$, ${\hat {\bm \alpha}}_t$ and ${\hat
{\bm \beta}}_t$ which are iteratively updated using Eq.~\ref{eq:krrupdate} ($\eta$ is the update rate).
\begin{equation}
\begin{array}{l}
{{{\bf{\hat D}}}_{t}} = (1- \eta) {{{\bf{\hat D}}}_{t-1}} + \eta{{\bf{D}}_{t-1}}\\
{{\hat {\bm \alpha} }_{t}} = (1-\eta) {{\hat {\bm \alpha} }_{t-1}} +  \eta{ {\bm \alpha} _{t-1}}\\
{{\hat {\bm \beta} }_{t }} = (1-\eta) {{\hat {\bm \beta} }_t} + \eta{{\bm \beta} _{t-1}}
\end{array}.
\label{eq:krrupdate}
\end{equation}

\begin{figure*}[http]
\centering
\begin{tabular}{c}
\includegraphics[width=1.0\linewidth,height=4.6cm]{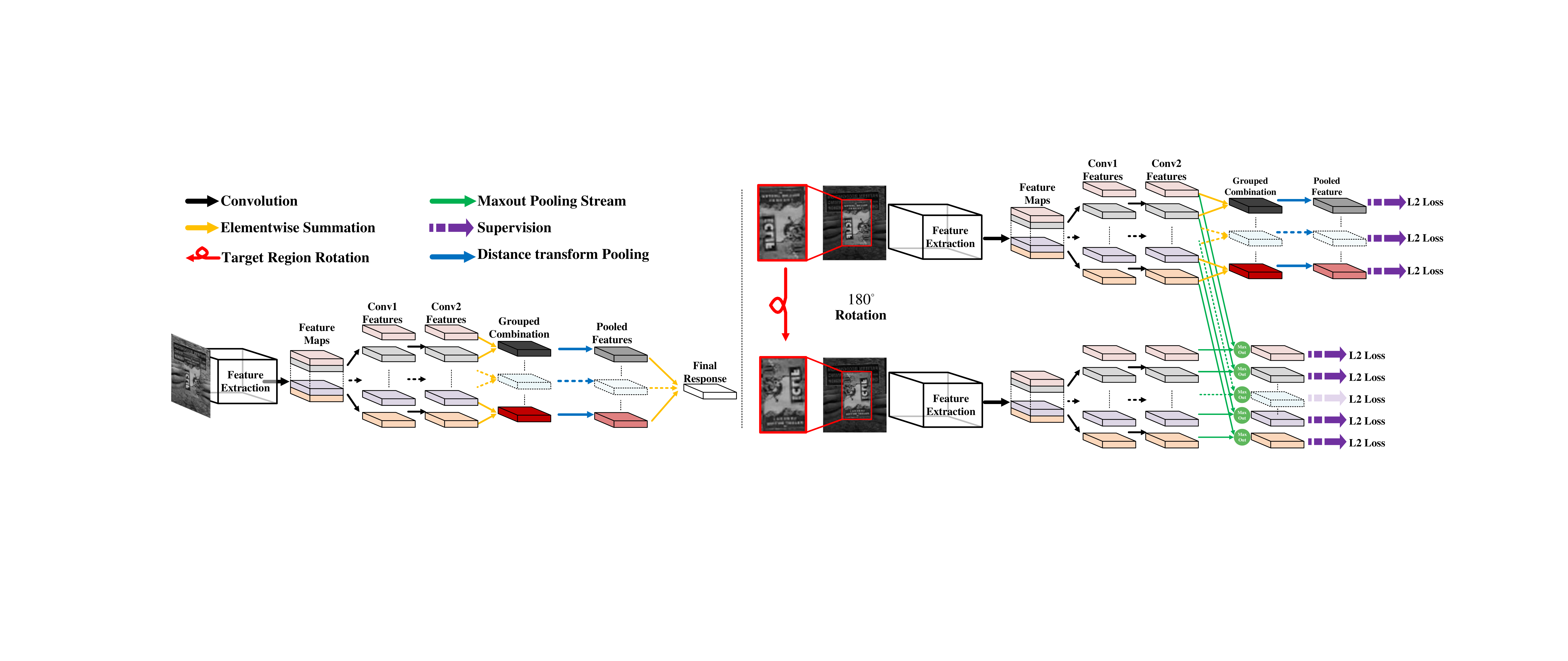}\\
\label{fig:network_structure}
\vspace{-4mm}\\\PutCaptrain
\end{tabular}
\caption{Network structures for our CNNSRK model.
(a) The testing network for CNNSRK. We use a convolutional neural network to
estimate the target position and exploit the distance transform pooling layer to
determine the effectiveness of each response map.
(b) The training network for CNNSRK. We utilize the two-stage training strategy
to update/train the convolution and distance transform pooling layers separately.
A two-stream network is used to learn the rotation information of the target.
\textbf{Best viewed in color with zoom in.}}
\label{fig:cnnsrk}
\end{figure*}

\section{Spatial-Aware CNN}
As is described in previous papers (\eg,~\cite{adam2006robust,kim2015sowp}), the spatial
information plays an important role in a visual tracking system. However, the spatial
layouts of the target object are usually ignored in the current CNN-based tracking algorithms.
What is more, the existing algorithms do not perform well when the target object suffers
from a severe in-plane rotation. In this work, we propose a convolution layer with spatially
regularized filter kernels, through which each convolution kernel only focuses on a specific
region of the target. In addition, as the training samples in visual tracking are very limited,
we develop a two-stream training strategy to avoid overfitting and consider rotation information.
\subsection{CNN with Spatially Regularized Kernels (CNNSRK)}
The proposed CNN framework with spatially regularized kernels is illustrated in Figure~\ref{fig:cnnsrk} (a),
which consists of two convolution layers interleaved with an ReLU layer and several distance transform
pooling layers.
Given the input feature map, we first reshape it to $46\times 46\times C$, where $C$ is the channel number.
Based on the input feature map, the first convolution layer (conv1) has a kernel size of $5\times 5$, and outputs a feature map with
size $46\times46\times C_1$.
The second convolution layer (conv2) has a kernel size of $3\times 3$ (we divide the input feature
into $C_1$ groups), and outputs a $46\times46\times C_1$ response map, each channel of which is a heat
map for the target localization.
After that, we divide the response map into several groups ($C_1/4$ groups in our implementation), and sum the
responses in each group through the channel dimension, which results in a $46\times 46\times 1$ output for
each group.
Finally, the obtained response maps are fed into distance transform pooling layers, and the outputs are
effectively combined to produce a final heat map for target localization.
The detailed structure is shown in Figure~\ref{fig:cnnsrk} (a).

\noindent{\textbf{Convolution Layer with Spatially Regularized Filter Kernels: }}
The target object may experience deformations and occlusions during tracking,
which makes some parts of the target object more important than others.
To address this issue, Wang~\etal~\cite{wang2016stct} modifies the implementation
of the Dropout layer and keep the dropped activations fixed in the training process.
By doing this, the learned convolution layers are forced to focus on different parts
of the input feature map.
But the discriminant ability of their model is weak as it merely takes parts of the
input feature map for consideration in the training process.
In this work, instead of introducing constraints on the input feature map, we enforce
constraints on the filter kernels in the convolution layer. Compared with \cite{wan2013regularization},
our method considers the spatial information, and fixes the filter mask during the tracking process.

Let ${\bf F}_c\in \mathbb{R}^{K_h\times K_w\times K_c}$ denotes the filter kernel weights
associated with the $c$-th channel of the output feature map ${\bf O}_c$. We introduce the
spatial regularization weights ${\bf W}_c$ into the convolution layer, which has the same
size as ${\bf F}_c$. By considering the spatial regularization weights, the output feature
map ${\bf O}_c$ can be calculated as,
\begin{equation}
  {\bf O}_c=({\bf F}_c\odot{\bf W}_c)*{\bf X}_c+b,
\end{equation}
where $*$ denotes the convolution operation, $\odot$ stands for the Hadamard product,
${\bf X}_c$ represents the input feature map and $b$ is the bias term.
For constructing ${\bf W}_c$, we first generate a binary mask ${\bf M}_c$ of size
$K_h\times K_w$ through Bernoulli distribution $B(0.3)$ and then construct ${\bf W}_c$
based on this mask,
\begin{equation}
 {{\bf{W}}_c}(p,q,r) = {{\bf{M}}_c}(p,q),
\end{equation}
where $p$, $q$ and $r$ denote the indexes of a 3-dimensional matrix. Clearly, only parts
of the spatial regions in ${\bf{W}}_c$ have non-zero values, which forces the filter kernels
to focus on different regions.

\vspace{1mm}
\noindent{\textbf{Distance Transform Pooling Layer: }} The distance transform has been used
in previous studies for object detection (\cite{felzenszwalb2010object,savalle2014deformable}).
Girshick~\etal~\cite{girshick2015deformable} point out that distance transform is indeed a
generalization of the max pooling layer and can be expressed in a similar formula as max pooling.

Given a function $y=f(x)$ defined on a regular grid $\mathcal{G}$, its distance transform can be
calculated as,
\begin{equation}
{D_f}(s) = \max_{{t \in \mathcal{G}}}(f(t) - d(s - t)).
\end{equation}
Here $d(s - t)$ is a convex quadratic function with $d(s - t)=\varpi (s-t)^2+\theta(s-t)$,
where $\varpi$ and $\theta$ are learnable parameters. The distance transform pooling layer
can be used to estimate the reliability of the input feature map. Generally speaking,
the larger the learned value $\varpi$ is, the more reliable the input feature map is.
When $\varpi=\theta=0$, this layer outputs a response with constant values, which means
that the input feature map does not influence the tracking result. In this work, the distance
transform pooling layer is implemented in the Caffe framework according
to~\cite{goodfellow2013maxout} and the pooling region is bounded for efficient computation.

\subsection{Two-Stream Training Strategy}
Based on our observation that the tracking performance is inevitably influenced
when severe in-plane-rotation\footnote{Rotation angle is larger than 90 degrees.}
occurs, we develop a two-stream network to learn the weights of convolution and
distance transform pooling layers. The proposed two-stream network is presented
in Figure~\ref{fig:cnnsrk} (b), in which the weights of the convolution layers are shared.

\begin{figure}[t]
\centering
\begin{tabular}{c@{}c}
\includegraphics[width=0.48\linewidth,height=24mm]{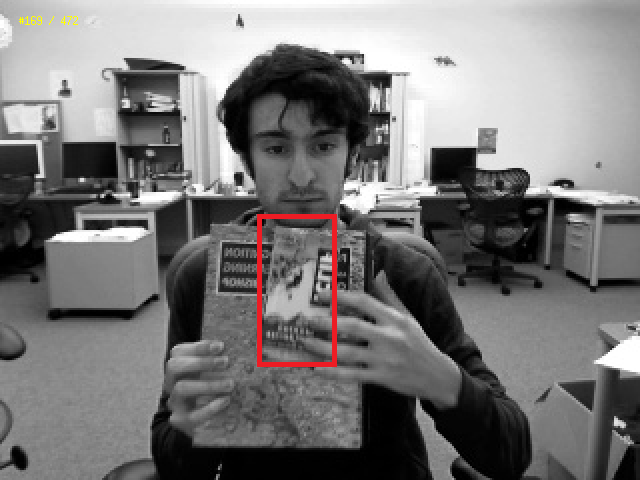}
\ &
\includegraphics[width=0.48\linewidth,height=24mm]{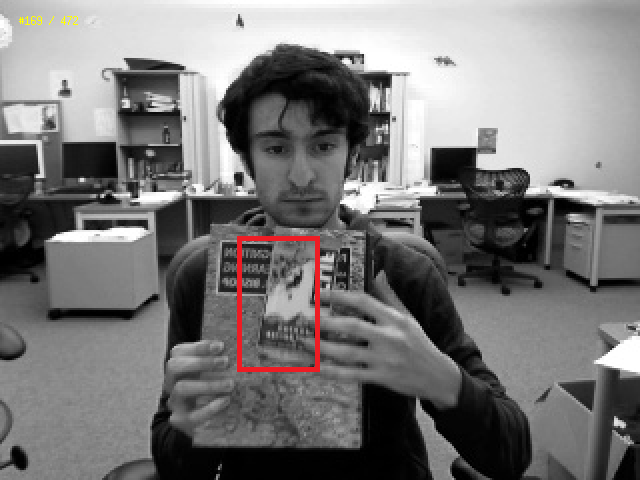}
\ \\
\includegraphics[width=0.48\linewidth,height=24mm]{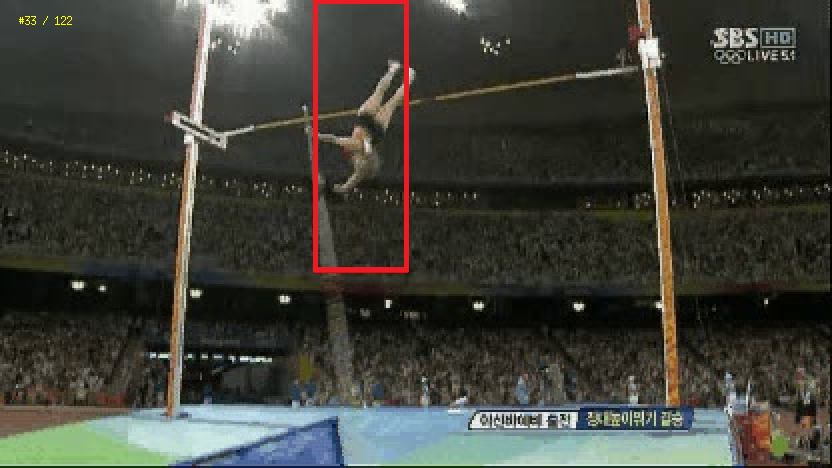}
\ &
\includegraphics[width=0.48\linewidth,height=24mm]{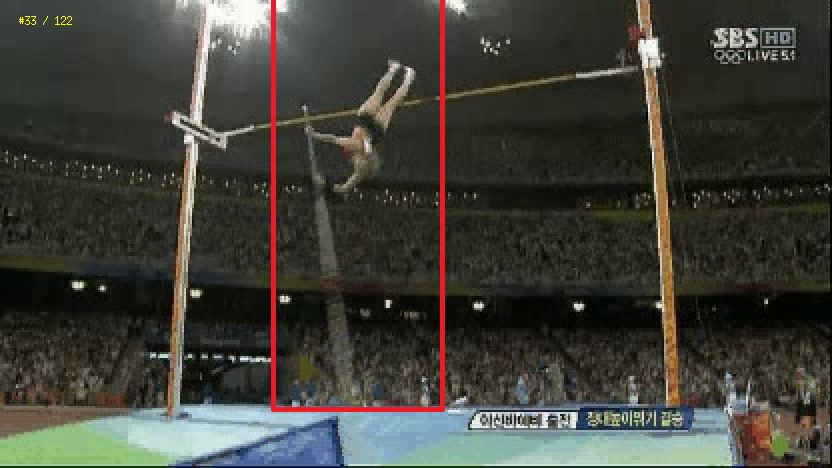}
\ \\
(a) & (b)
\end{tabular}
\caption{Tracking results with and without two-stream training
process are illustrated in (a) and (b), respectively. We can see
that the proposed two-stream training strategy makes the tracker
perform better when severe in-plane rotation occurs. }
\label{fig:rotation_result}
\end{figure}

The upper branch of the network exploits the same feature map as the original
network (Figure~\ref{fig:cnnsrk} (a)), and the lower branch uses the input
feature map corresponding to the rotated target object.
The previous two layers are the same as the original network, resulting in two
$46\times46\times C_1$ response maps in both upper and lower branches.
Then, we conduct the max-out pooling operation on these two response maps to
produce a $46\times46\times C_1$ response.
We compute the loss for each channel of the response maps and propagate the
loss backward to learn the filters in convolution layers.
After that, we fix the convolution layers and learn the model parameters of
the distance transform pooling layers.
Some visual results are illustrated in Figure~\ref{fig:rotation_result},
from which we can see that the proposed two-stream training strategy
facilitates dealing with severe in-plane rotation.

In addition, the reason why we do not directly perform model learning on
the original network (presented in Figure~\ref{fig:cnnsrk} (a)) is to avoid
the overfitting problem. For example, the original network has a total of
$5\times5\times C\times C_1+3\times3\times C_1+C_1$
parameters. If $C=512$ and $C_1=100$, there will be a total of
1281000 parameters, which are very difficult to be learned with limited training
data in visual tracking. The proposed two-stream strategy essentially
decomposes the network into several sub-parts and trains
each part separately, thereby facilitating avoiding overfitting.

\section{Tracking with Spatial-Aware KRR and CNN}
In this section, we describe how to exploit both spatial-aware
KRR and CNN models for robust visual tracking.
\subsection{Target Location Estimation}
We conduct visual tracking by combining the responses of our KRR and CNN models. The former one
captures the holistic information of the target, while the latter one focuses more on the localized region.
In frame $t$, we crop a search region centered by the estimated object location of the last frame
and then obtain a feature map ${\bf X}_t$ for this region.
Furthermore, the final heat map of our tracker can be computed as,
\begin{equation}
{\bf f}({{\bf{X}}_t}) = {{\bf f}_{KRR}}({{\bf{X}}_t}) + \gamma{{\bf f}_{CNN}}({{\bf{X}}_t}),
\end{equation}
where $\gamma$ is a trade-off parameter. ${{\bf f}_{KRR}}({{\bf{X}}_t})$ and ${{\bf f}_{CNN}}({{\bf{X}}_t})$
denote the heat maps produced by KRR and CNN models, respectively.
Finally, we find the position with highest heat map score in ${\bf f}({{\bf{X}}_t})$ and determine it as the
optimal location.
\subsection{Scale Estimation}
It is not enough to only provide the location for a target object, which may experience drastic scale variation.
In this work, we also estimate the scale variation of the target after location estimation.
We use $S$ to denote the candidate scale size and $H\times W$ to denote the input feature map size.
For each ${s_l} \in \left\{ {\left\lfloor { - \frac{{S - 1}}{2}} \right\rfloor ,...,\left\lfloor {\frac{{S - 1}}{2}}
\right\rfloor } \right\}$, we crop or pad the input feature map to size $a^{s_l}H\times a^{s_l}W$ and reshape
it to $H\times W$ (we use ${\cal T}({\bf X}_t,s_l)$ to denote the transformed feature map for scale $l$).
These transformed maps are then fed into a fully connected layer to output the scale scores. Finally, we choose
the scale related to the largest score as the optimal state. After scale estimation, we further exploit the bounding
box regression method~\cite{girshick2014rich,nam2016learning} to refine the tracking result.

\subsection{Model Update}
After obtaining the optimal location and scale, we crop the corresponding
search region and extract its feature map ${\bf X}_t$. Then, we generate
an ideal heat map of Gaussian distribution~\cite{henriques2015high} and
exploit the L2 loss function for finetuning both KRR and CNN models to
fit the ideal heat map. In this work, the stochastic gradient descent
(SGD) method is adopted for finetuning both networks.
For KRR, after model parameters ${\bm \alpha}_t$ and ${\bm \beta}_t$ are updated, we further update
${\bf \hat D}_t$, ${\hat {\bm \alpha}}_t$ and ${\hat {\bm \beta}}_t$ based on Eq.~\ref{eq:krrupdate} .
For CNN, our two-stream network (Section 4.2) is adopted for updating model parameters.

In addition, if scale variation is detected, we update the scale estimation network with the loss function ${{\cal L}_{\cal S}}
= \frac{1}{2}\left\| {{y}_{s_l} - {\bf{f}}_{\cal S}({\cal T}({{\bf{X}}_t},{s_l}))} \right\|_2^2$ (${\bf{f}}_{\cal S}$ denotes
the score obtained by the network, ${y_{{s_l}}} = \exp ( - \frac{1}{{2{\sigma ^2}}}s_l^2)$ is a Gaussian function).

\vspace{1mm}
\section{Experimental Results}
In this section, we first introduce the experimental setups, and then report the experimental
results of different trackers on the OTB-2015 and VOT-2017 public datasets.
In addition, we verify the effectiveness of different components of the proposed method.
Our source codes can be downloaded at~\url{https://github.com/cswaynecool/LSART}.

\subsection{Implementation Details}
The proposed tracker is implemented with MATLAB2014a on an Intel 4.0 GHz CPU
with 32G memory, and runs at around 1fps during online tracking.
We use the Caffe toolbox~\cite{jia2014caffe} to implement the networks, whose
forward and backward operations are conducted on a Nvidia Titan X GPU.
We divide the target into 9 ($3\times 3$) patches in the kernelized ridge
regression model. The trade-off parameters $\lambda_1$, $\lambda_2$
are set as 0.001 and 0.001 respectively.
The learning rate $\eta$ is set as $0.2$ in the first ten frames, and changed
to a smaller learning rate (\eg $0.001$) during the tracking process.
All the networks are trained with the SGD method, and the learning rates for
${\bm \alpha}_t$ and ${\bm \beta}_t$ in the KRR network are set as 8e-9 and
1.6, while the learning rate for each layer in the CNN network is fixed as
8e-7.

\subsection{OTB-2015 Dataset}
The OTB-2015 dataset~\cite{wu2015object} is one of the most commonly used
benchmarks in evaluating different trackers. This dataset includes $100$ challenging image
sequences with 11 different attributes, such as illumination variation, background clutter,
scale variation, fast motion, in-plane rotation and so on.
In this dataset, we exploit the the output of the Conv4-3 layer
of the VGG-16 net as the basic feature for our tracker, and compare our LSART method with 12 state-of-the-art trackers including DSST~\cite{danelljan2014accurate},
ECO~\cite{danelljan2016eco}, CCOT~\cite{danelljan2016beyond}, SRDCF~\cite{danelljan2015learning}, KCF~\cite{henriques2015high},
DeepSRDCF~\cite{danelljan2015convolutional}, CF2~\cite{ma2015hierarchical}, LCT~\cite{ma2015long},
SRDCFdecon~\cite{danelljan2016adaptive},  HDT~\cite{qi2016hedged},  Staple~\cite{bertinetto2016staple} and MEEM~\cite{zhang2014meem}.
We exploit the one-pass evaluation (OPE) for all the trackers and report both the precision and success plots for comparison.
The precision plots aim to measure the percentage of frames in which the distance between the tracked result and the ground-truth
is under a threshold, while the success plots aim to measure the successfully tracked frames with various thresholds.
Following~\cite{WuLimYang13}, in the precision plots, we use the distance precision rate at threshold 20 for ranking,
while in the success plots, we use the area under curve (AUC) for ranking.

\begin{figure}[t]
\centering
\begin{tabular}{c@{}c}
\includegraphics[width=0.5\linewidth,height=31mm]{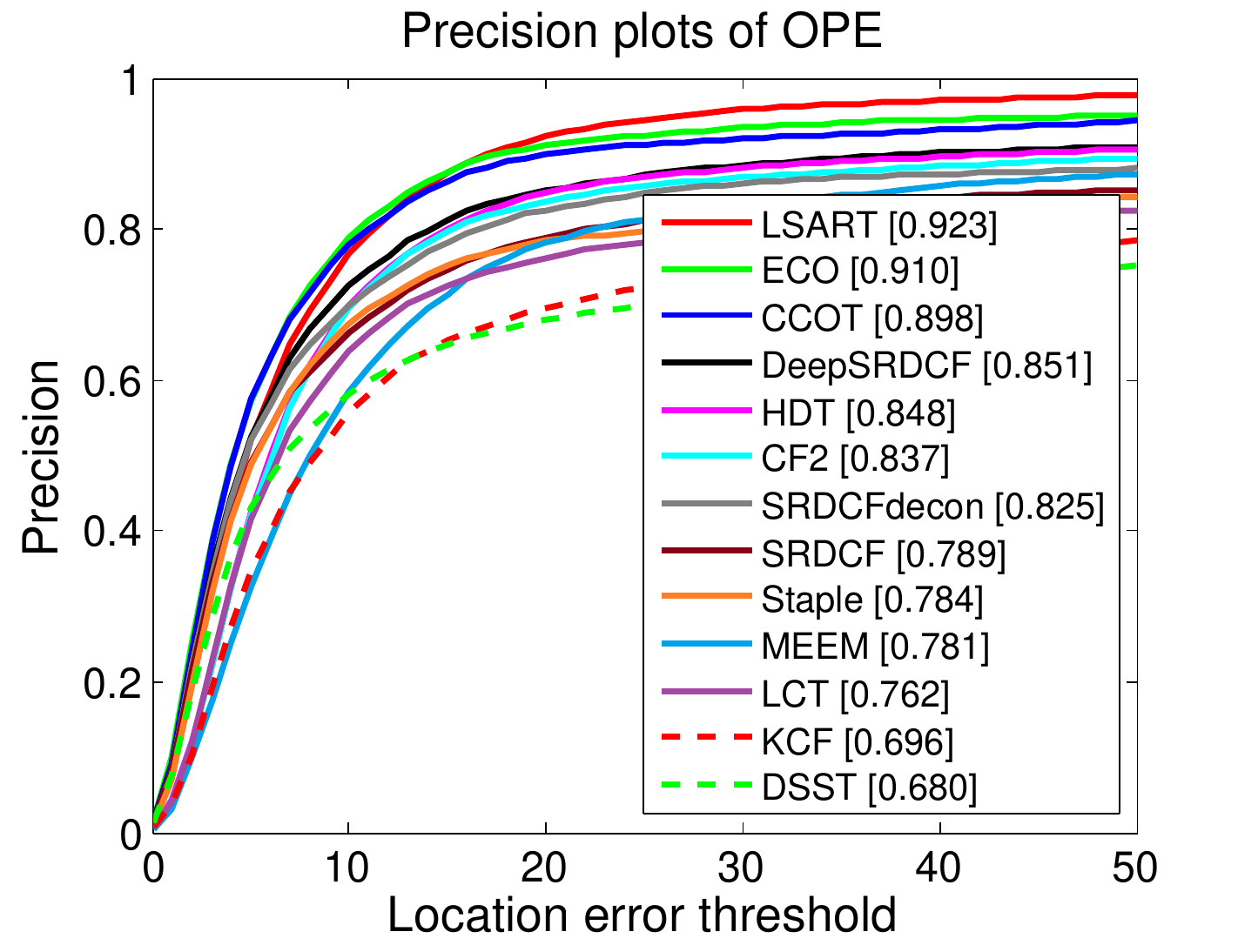}
\ &
\includegraphics[width=0.5\linewidth,height=31mm]{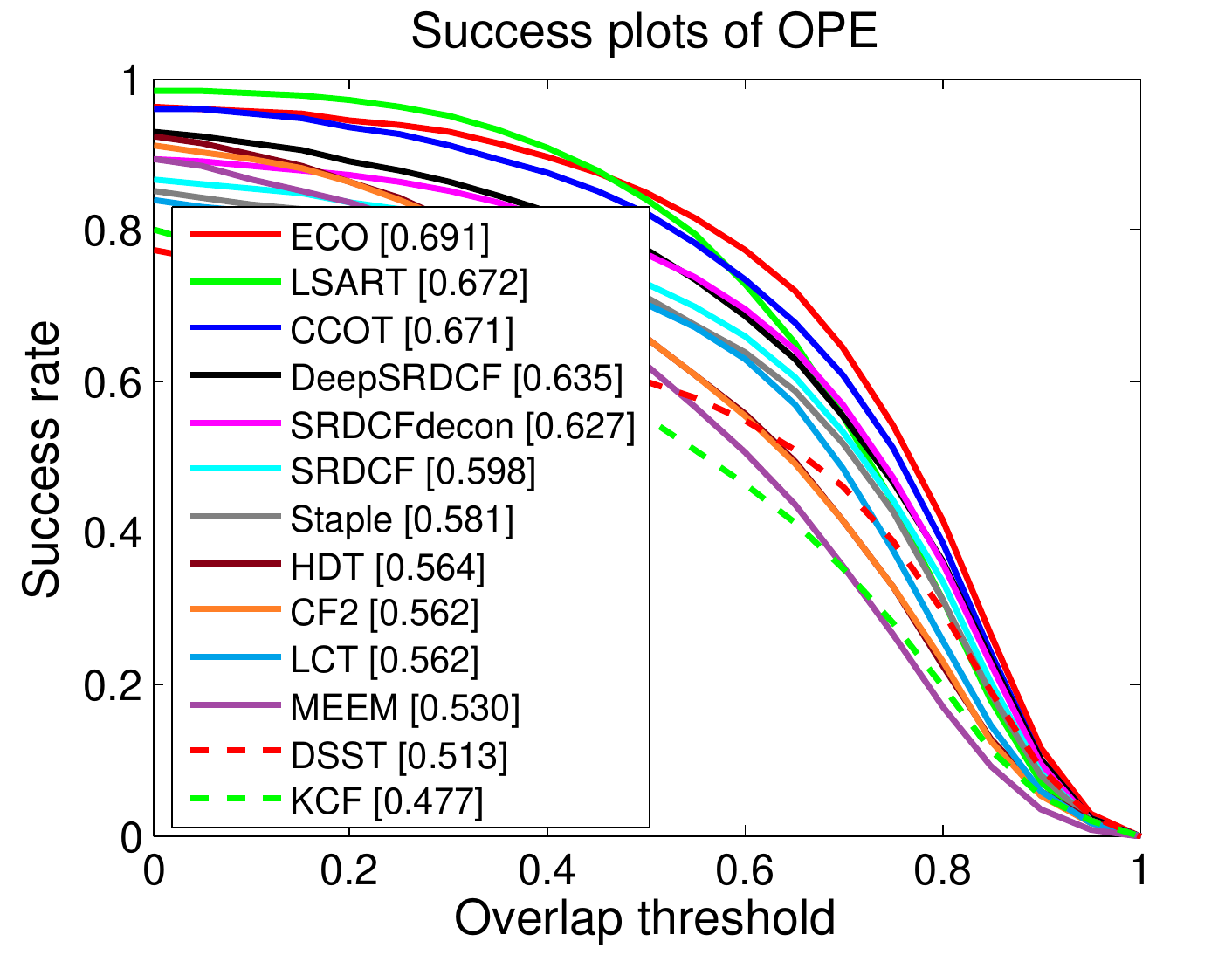}
\ \\
\end{tabular}
\caption{Precision and success plots on the OTB-2015 dataset in terms of OPE rule.
In the legend, we show the distance precision rates at threshold 20 and area under curve (AUC) scores,
based on which different trackers are ranked.}
\label{fig:otb-2015}
\vspace{-2mm}
\end{figure}

Figure~\ref{fig:otb-2015} illustrates both precision and success plots over all $100$ videos in this dataset.
We can see that our LSART method achieves the best performance with a precision score $92.3\%$
and the second best result with an AUC score of $67.2\%$.
Overall, in OTB-2015, our tracker has comparable performance compared to the existing best tracker
ECO.

\begin{figure*}[t]
\centering
\begin{tabular}{c@{}c@{}c@{}c}
\includegraphics[width=0.21\linewidth,height=26mm]{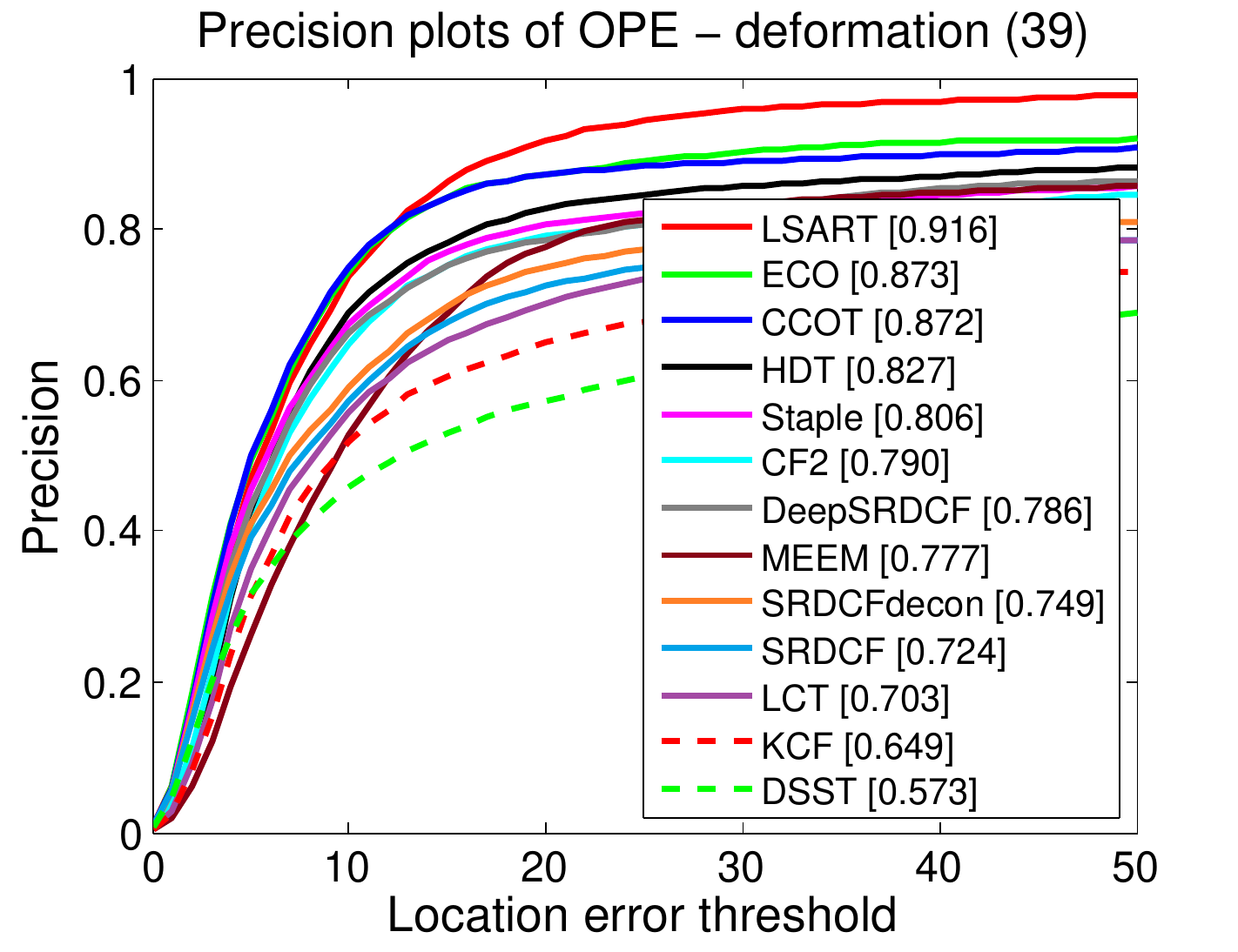}
\ &
\includegraphics[width=0.21\linewidth,height=26mm]{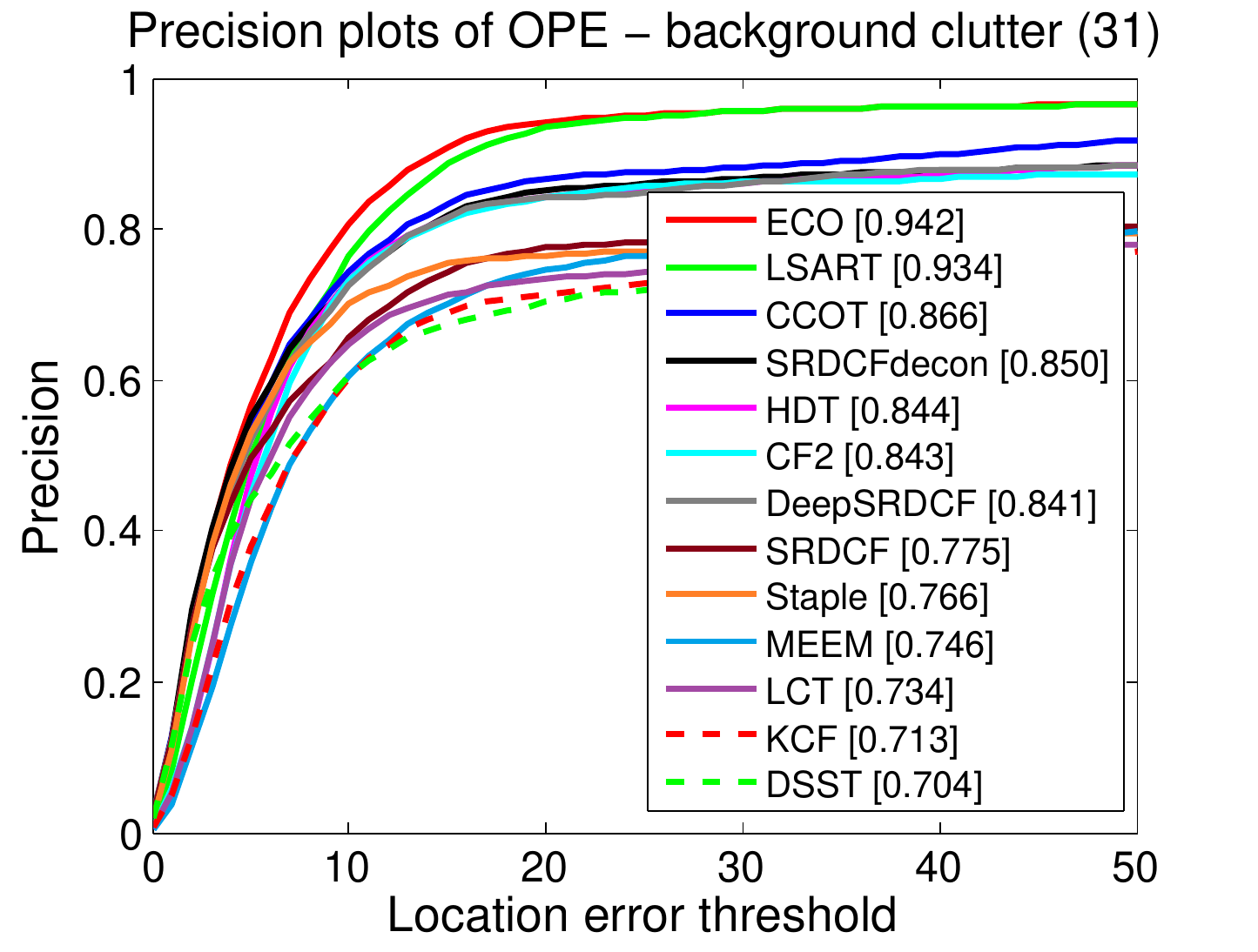}
\ &
\includegraphics[width=0.21\linewidth,height=26mm]{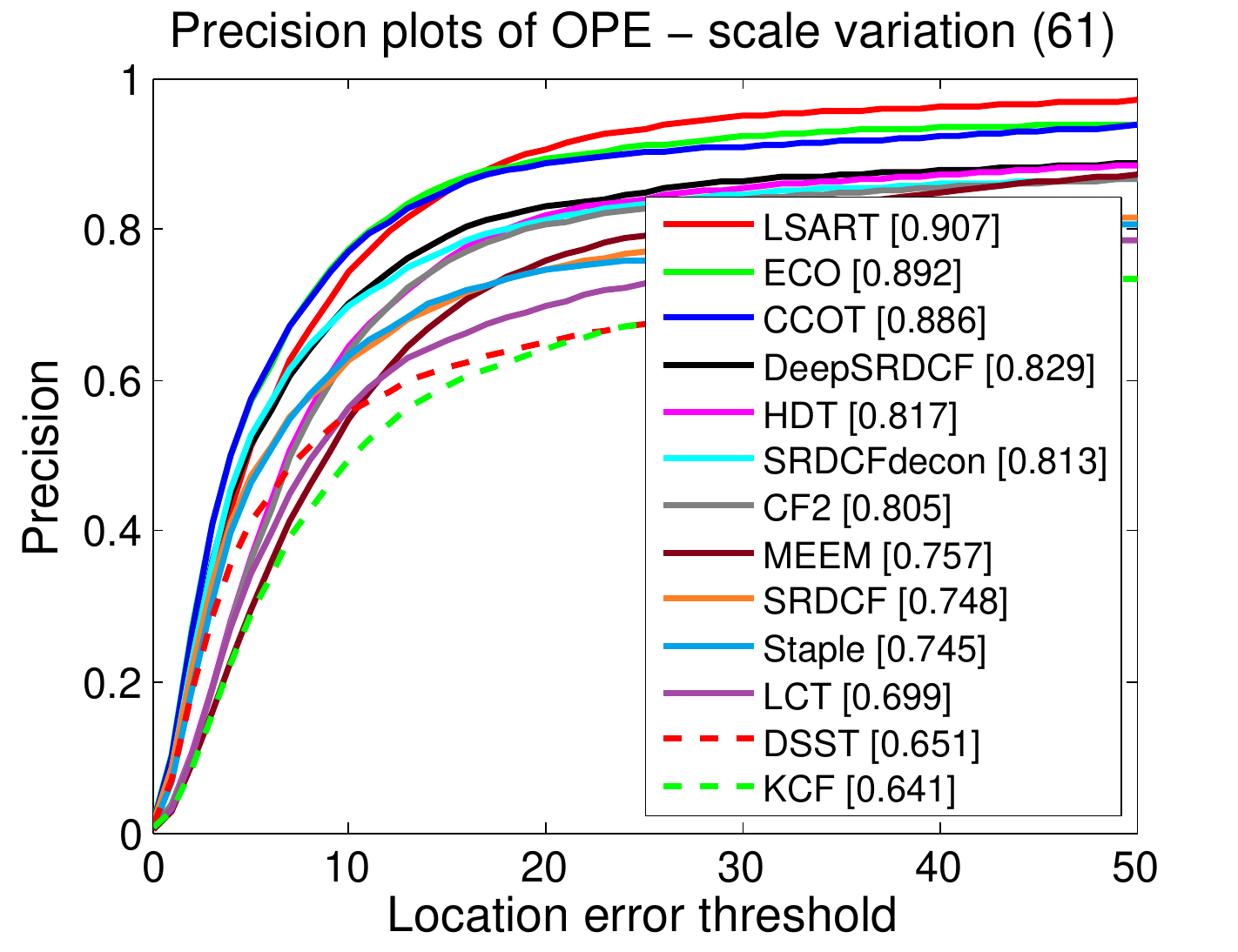}
\ &
\includegraphics[width=0.21\linewidth,height=26mm]{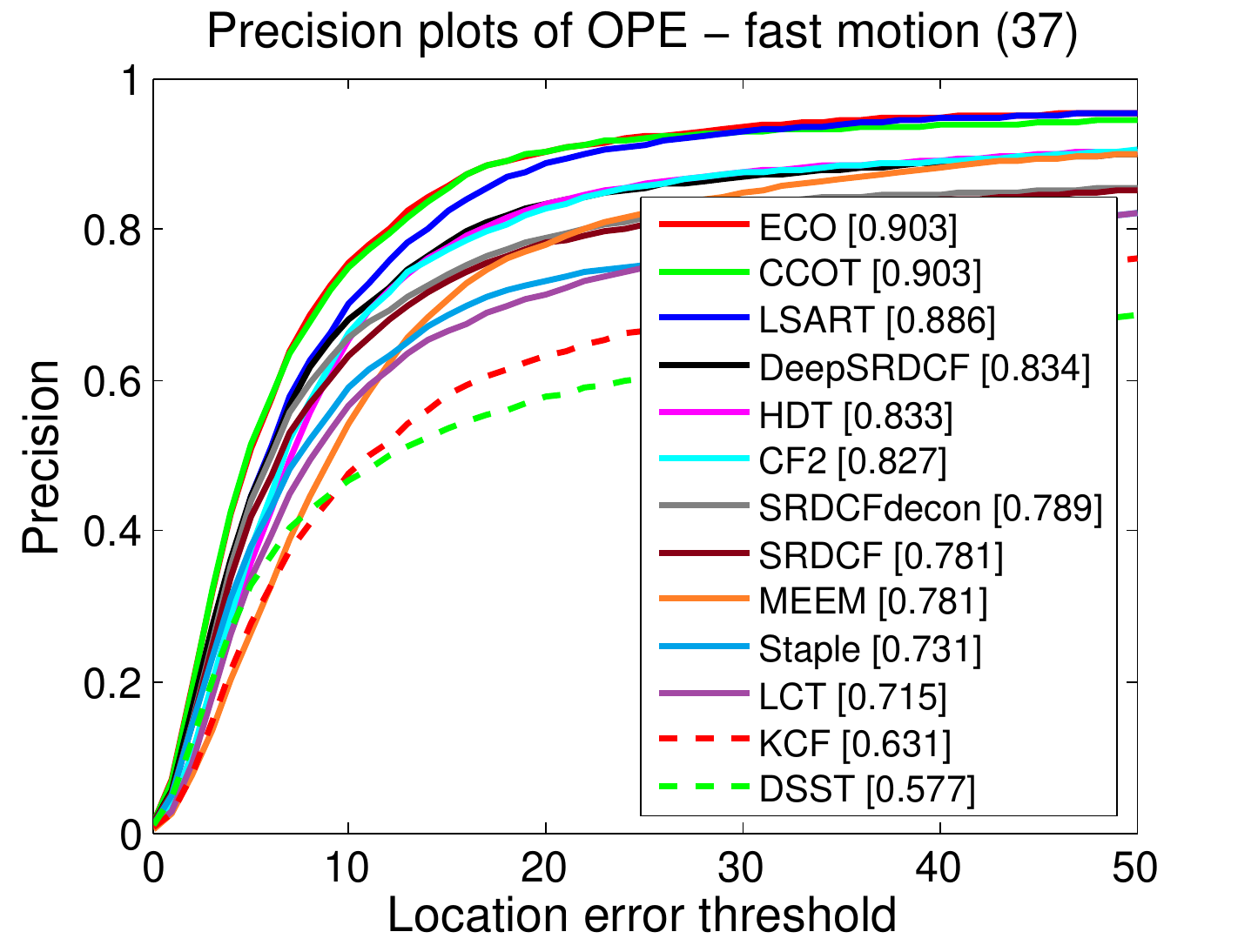}
\ \\
\includegraphics[width=0.21\linewidth,height=26mm]{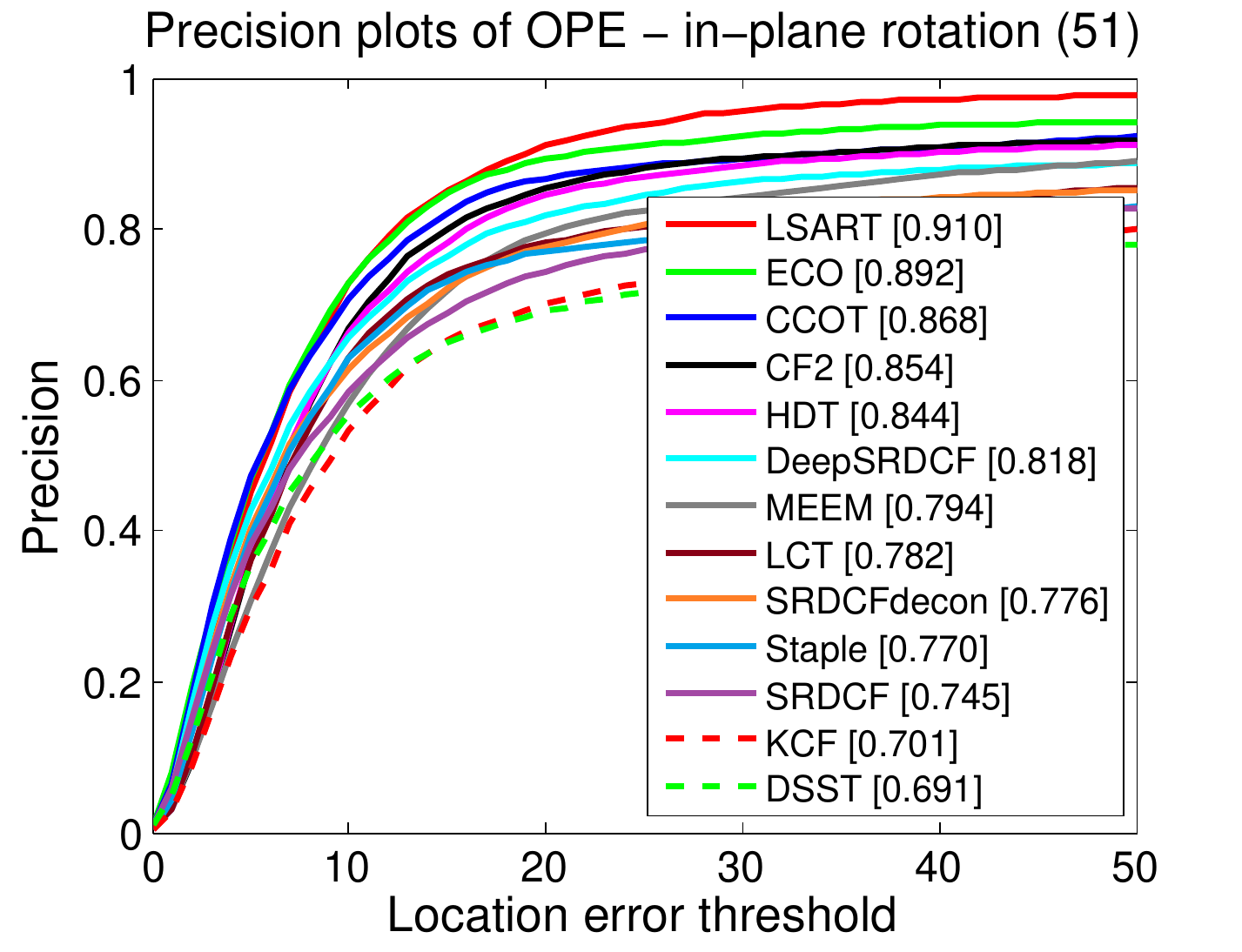}
\ &
\includegraphics[width=0.21\linewidth,height=26mm]{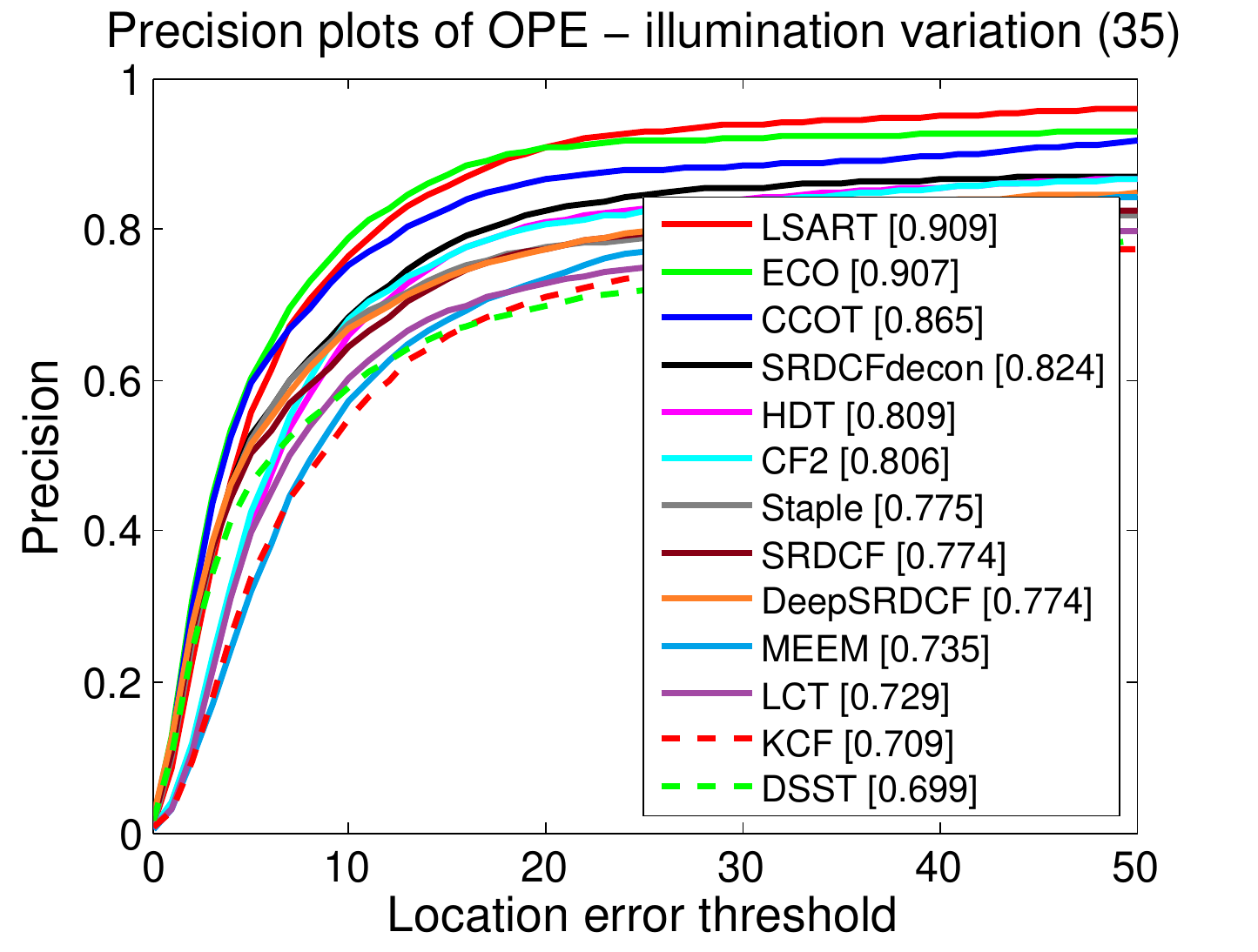}
\ &
\includegraphics[width=0.21\linewidth,height=26mm]{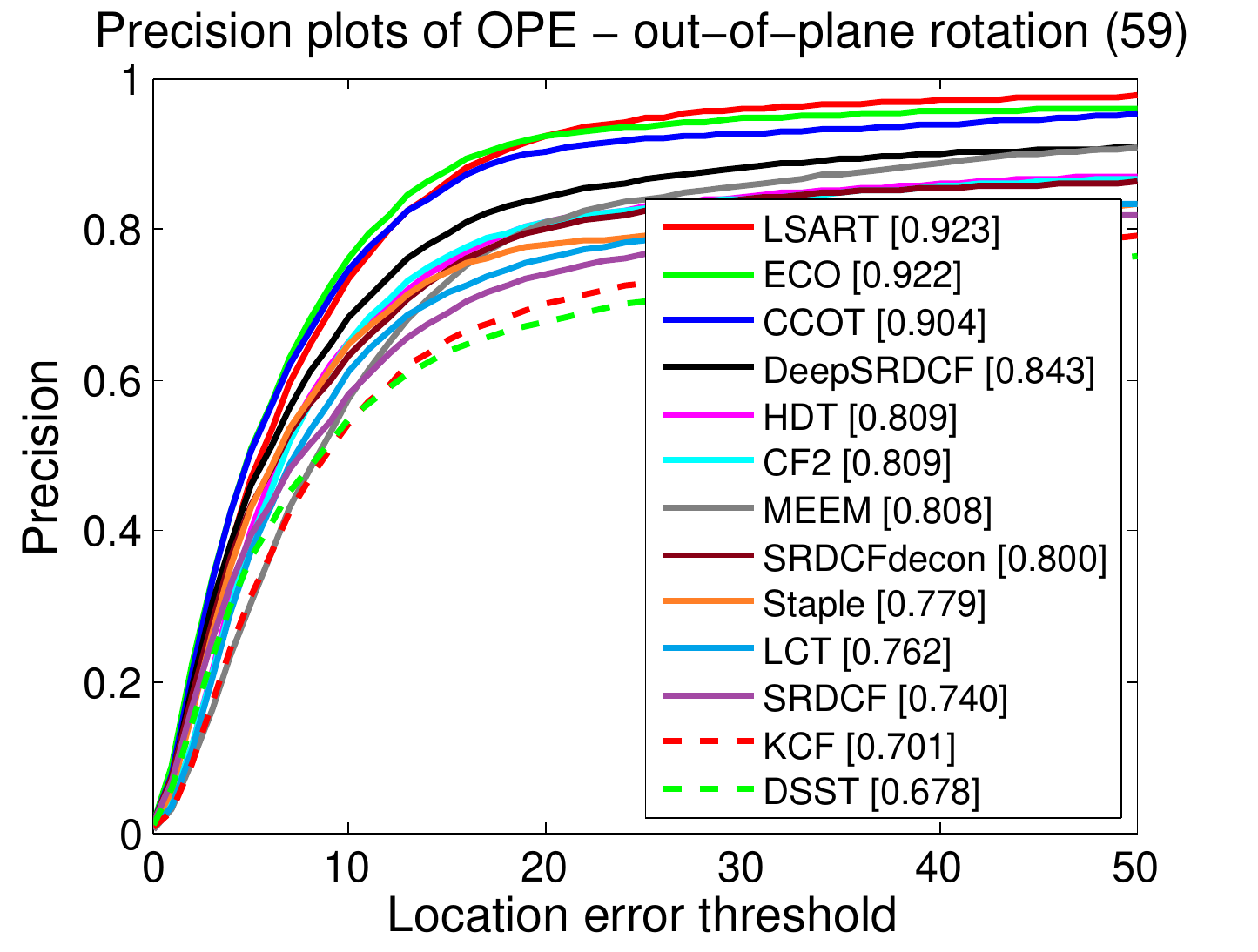}
\ &
\includegraphics[width=0.21\linewidth,height=26mm]{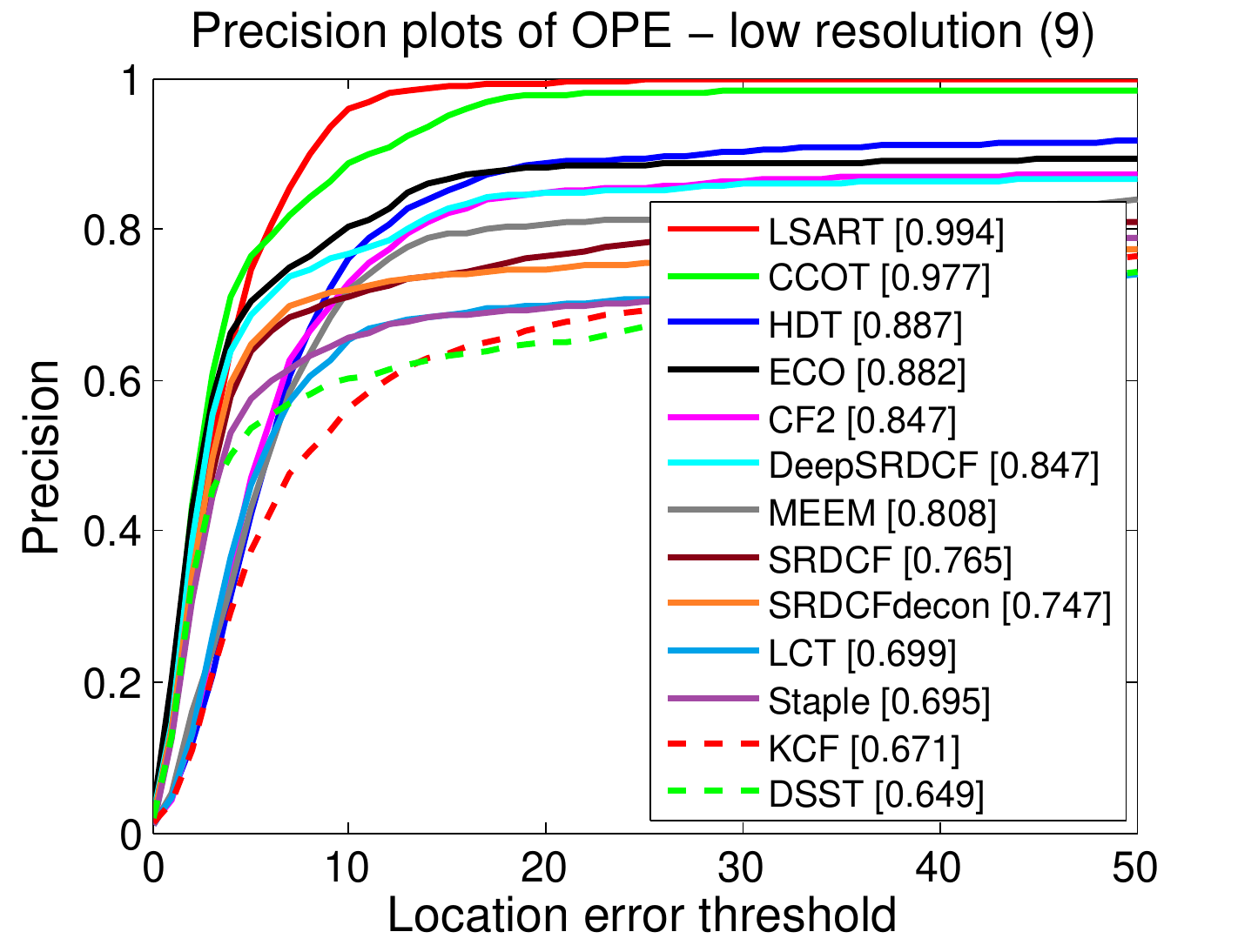}

\end{tabular}
\caption{Performance evaluation on different attributes of OTB-2015 in terms of the OPE criterion.
The reported attributes include deformation, background clutter, scale variation, fast motion,
in-plane rotation, illumination variation, out-of-plane rotation, low resolution.}
\label{fig:attributes}
\vspace{-2mm}
\end{figure*}

Besides, we evaluate different trackers using 8 attributes and report their precision plots
in Figure~\ref{fig:attributes}.
The results show that our tracker achieves very promising performance in handling most of the challenges,
especially for deformation and in-plane rotation.
First, the proposed LSART method improves the second best one by 4.3\% in the attribute of deformation.
This improvement is mainly because that our tracker determines the reliability of the target object adaptively
and is insusceptible to the unreliable regions.
In addition, for in-plane rotation, our method improves the ECO method by 1.8\%. The underlying reason is that
our tracker exploits the two-stream network to learn the rotation information of the target object effectively.

\begin{table}[hht]
\caption{Performance evaluation for 12 state-of-the-art algorithms on the VOT-2017 public dataset.
The best three results are marked in red, blue and green bold fonts respectively.}
\label{tab:table1}
\small
\begin{center}
\label{tab1}
\begin{tabular}{cccccc}
\hline
Tracker&EAO&A&R&AO
\\
\hline
\hline
LSART&\color{red}{\bf 0.323}&0.493&\color{red}{\bf 0.218}&\color{blue}{\bf 0.437}\\
CFCF&\color{blue}{\bf 0.286}&\color{green}{\bf 0.509}&\color{green}{\bf 0.281}&0.380\\
ECO&\color{green}{\bf 0.280}&0.483&\color{blue}{\bf 0.276}&\color{green}{\bf 0.402}\\
CCOT&0.267&0.494&0.318&0.390\\
MCPF&0.248&\color{blue}{\bf 0.510}&0.427&\color{red}{\bf 0.443}\\
CRT&0.244&0.463&0.337&0.370\\
ECOhc&0.238&0.494&0.435&0.335\\
MEEM&0.192&0.463&0.534&0.328\\
FSTC&0.188&0.480&0.534&0.334\\
Staple&0.169&\color{red}{\bf 0.530}&0.688&0.335\\
KCF&0.135&0.447&0.773&0.267\\
SRDCF&0.119&0.490&0.974&0.246&\\
\hline
\vspace{-4mm}
\label{tab:vot-2017}
\end{tabular}
\end{center}
\end{table}

\vspace{-0mm}
\subsection{VOT-2017 Public Dataset}
For more thorough evaluations, we test our LSART tracker on the VOT-2017 public dataset~\cite{VOT2017}
in comparison with 11 state-of-the-art methods, including ECO~\cite{danelljan2016eco},
CCOT~\cite{danelljan2016beyond}, CFCF~\cite{gundogdu2017good}, MCPF~\cite{zhang2017multi},
CRT~\cite{chen2016convolutional}, ECOhc~\cite{danelljan2016eco}, MEEM~\cite{zhang2014meem},
FSTC~\cite{wang2015visual}, Staple~\cite{bertinetto2016staple}, KCF~\cite{henriques2015high}
and SRDCF~\cite{danelljan2015learning}.
Since many top-ranked trackers (\eg CFCF, ECO and CCOT) exploit a combination of CNN and hand-crafted features,
we extend our feature set with HOG and Color Naming like CCOT in this dataset.
The VOT-2017 public dataset is one of the most recent datasets for evaluating online model-free single-object trackers,
and includes $60$ public image sequences with different challenging factors.
Following the evaluation protocol of VOT-2017~\cite{VOT2017}, we adopt the expected average
overlap (EAO), accuracy and robustness raw values (A, R) and no-reset-based average overlap
(AO) to compare different trackers.
The detailed comparisons are reported in Table~\ref{tab:vot-2017}.

From Table~\ref{tab:vot-2017}, we can conclude that the proposed LSART
method achieves the top-ranked performance in terms of EAO, R and AO criteria
while maintaining a competitive accuracy.
Especially, our LSART tracker has the best performance among the compared trackers in terms of
the EAO measure, which is the most important metric on the VOT dataset.
Compared with the second best tracker (CFCF), the proposed method achieves a relative
performance gain of 12.94\%.
In addition, our tracker achieves a  substantial improvement over the popular ECO method,
with a relative gain of 15.36\% in EAO.
Note that our tracker has an EAO of 0.275 without the hand-crafted features, which still outperforms
CCOT and is competitive among the compared trackers.
The OPE rule is also adopted to evaluate different trackers and the AO values are reported to
demonstrate their performance.
From the last column in Table~\ref{tab:vot-2017}, we can see that our method achieves comparable
performance compared to the MCPF tracker and improves the ECO method by a relative gain of 8.71\%.

\begin{figure}[h]
\centering
\begin{tabular}{c@{}c}
\includegraphics[width=0.45\linewidth,height=29mm]{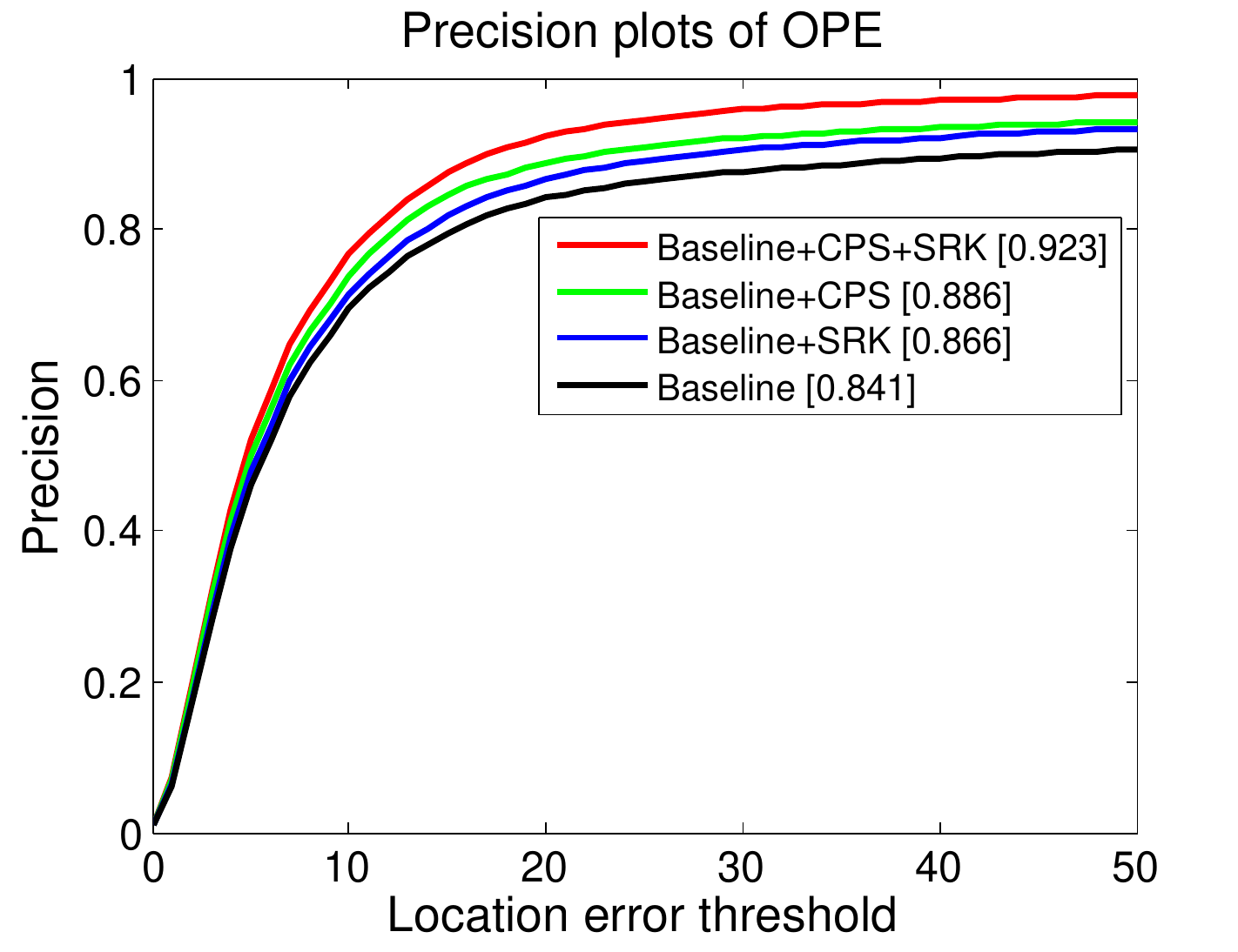}
\ &
\includegraphics[width=0.45\linewidth,height=29mm]{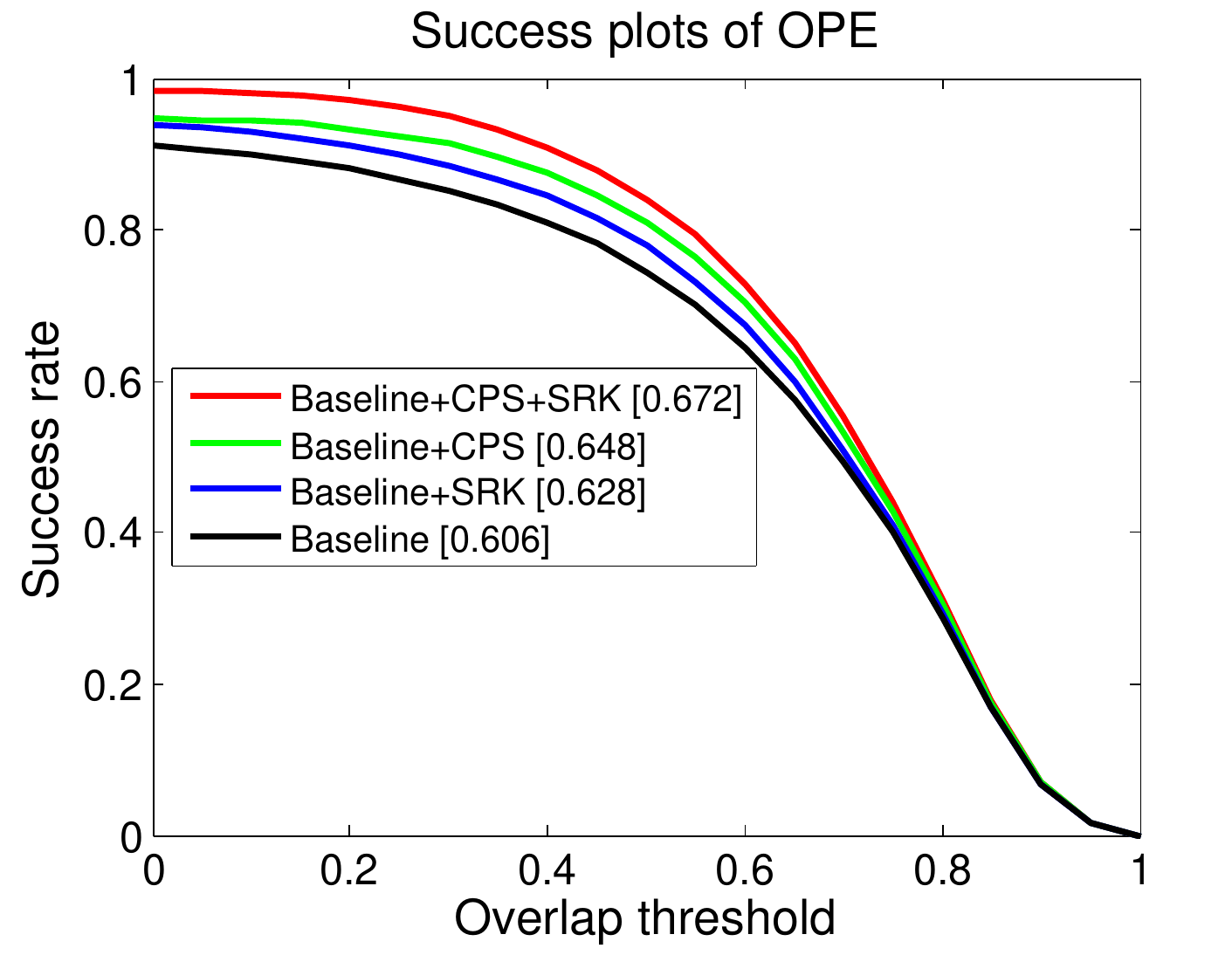}
\ \\
\end{tabular}
\caption{Performance evaluation for each component of the proposed tracker. }
\label{fig:analysis}
\vspace{-2mm}
\end{figure}

\vspace{1.05mm}
\subsection{Ablation Studies}
\label{sec:6-5}
In this paper, we propose two complementary spatial aware regressions for visual tracking, which are respectively
the kernelized ridge regression model with cross-patch similarity and convolution neural network with the spatially
regularized kernels.
Here, we conduct ablation analysis to evaluate each component of our tracker. With different experimental settings,
we obtain the following 4 variants of our tracker, which are respectively
named as ``Baseline'', ``Baseline+CPS'', ``Baseline+SRK'' and ``Baseline+CPS+SRK'' (LSART).
We use the shorthand ``Baseline'' to denote the method that directly combines the conventional KRR and  CNN models,
and adopt the abbreviations ``CPS'', ``SRK'' to denote the cross-patch similarity kernel and spatial regularized filter kernel.
Using OTB-2015, the results of different variants are presented in Figure~\ref{fig:analysis}.

First,  the direct combination of conventional KRR and CNN models (\ie, ``Baseline'') cannot achieve
satisfactory performance ($0.841$ in precision score and $0.606$ in AUC score).
Second, the effectiveness of the CPS module can be verified comparing``Baseline+CPS'' with ``Baseline'',
which contributes to the relative performance gains of $5.35\%$ and $6.93\%$ in precision and success plots.
The effectiveness of the SRK module can be validated by comparing ``Baseline+SRK'' with ``Baseline''.
Finally, we can see that our LSART method (``Baseline+CPS+SRK'') improves the
original ``Baseline'' method by relative gains of $9.75\%$ in precision plots and $10.89\%$ in success plots.

\vspace{-2mm}
\section{Conclusion}
This paper proposes a robust online tracker by exploiting both spatial-aware KRR and spatial-aware
CNN.
First, we propose a novel KRR model with cross-patch similarity (CPS). This model considers the interior
structure of the target, and can adaptively determine the importance of the similarity score between two
patches. We show that the KRRCPS model can be reformulated as a neural network, and thus can be more
efficiently solved.
In addition, we propose a complementary CNN model which focuses more on the localized region via
a spatially regularized filter kernel. Distance transform pooling layers are further exploited to determine
the reliability of the convolutional layers.
Finally, the above-mentioned two models are effectively combined to generate a final heat map for target
location. Experimental results on two recent benchmarks demonstrate that the proposed LSART method
achieves very promising tracking performance, especially on the VOT-2017 public dataset.

\textbf{Acknowledgment.} This paper is partially supported by the Natural Science
Foundation of China \#61725202, \#61502070, \#61472060, NSF CAREER (No.~1149783), gifts from Adobe, Toyota, Panasonic, Samsung, NEC, Verisk and Nvidia.
Chong Sun is also supported by the China Scholarship Council (CSC).

{\small
\bibliographystyle{ieee}

}

\end{document}